\documentclass{article}
\usepackage[letterpaper,top=2cm,bottom=2cm,left=3cm,right=3cm,marginparwidth=1.75cm]{geometry}
\usepackage{graphicx}
\usepackage{amsmath}
\usepackage{graphicx}
\PassOptionsToPackage{hyphens}{url}
\usepackage[colorlinks=true, allcolors=blue]{hyperref}
\usepackage{makecell}
\usepackage{booktabs}
\usepackage{tabularx}
\usepackage{ltablex}
\usepackage{caption}
\usepackage{subcaption}
\usepackage{stackengine}
\usepackage{multirow}
\usepackage{authblk}
\usepackage[table]{xcolor}
\usepackage{abstract}
\usepackage{breakcites}

\title{MIMDE: Exploring the Use of Synthetic vs Human Data for Evaluating Multi-Insight Multi-Document Extraction Tasks}

\author[1,\footnote{Corresponding author: \href{jfrancis@turing.ac.uk}{jfrancis@turing.ac.uk}}]{John Francis}
\author[1]{Saba Esnaashari}
\author[1]{Anton Poletaev}
\author[1,2]{Sukankana Chakraborty}
\author[1]{Youmna Hashem}
\author[1]{Jonathan Bright}

\affil[1]{Public Policy Programme, The Alan Turing Institute, British Library, UK.}
\affil[2]{Centre for Advanced Spatial Analysis, University College London, UK.}

\date{}
\begin{document}

\maketitle


\begin{abstract}
Large language models (LLMs) have demonstrated remarkable capabilities in text analysis tasks, yet their evaluation on complex, real-world applications remains challenging. We define a set of tasks, Multi-Insight Multi-Document Extraction (MIMDE) tasks, which involves extracting an optimal set of insights from a document corpus and mapping these insights back to their source documents. This task is fundamental to many practical applications, from analyzing survey responses to processing medical records, where identifying and tracing key insights across documents is crucial. We develop an evaluation framework for MIMDE and introduce a novel set of complementary human and synthetic datasets to examine the potential of synthetic data for LLM evaluation. After establishing optimal metrics for comparing extracted insights, we benchmark 20 state-of-the-art LLMs on both datasets. Our analysis reveals a strong correlation (0.71) between the ability of LLMs to extracts insights on our two datasets but synthetic data fails to capture the complexity of document-level analysis. These findings offer crucial guidance for the use of synthetic data in evaluating text analysis systems, highlighting both its potential and limitations.
\end{abstract}

\section{Introduction}
\label{sec:introduction}
Since the release of ChatGPT in late 2022, research into large language models (LLMs) has seen unprecedented growth and advancement which has also driven significant progress in the field of natural language processing (NLP). This has ultimately led to a remarkable increase in the types of tasks that can be automated using LLMs across various domains. While a large fraction of new tasks afforded by LLMs involve the generation of novel text data \cite{ray2023chatgpt}, they have also demonstrated high-quality performance on a variety of traditional NLP tasks such as sentiment analysis \cite{zhang2023sentiment}, summarisation \cite{pu2023summarization}, text classification \cite{chae2023large}, and textual similarity \cite{gatto2023text}. In this paper, we argue that LLMs have also opened up opportunities for new tasks within NLP which were previously not possible with traditional text analysis tools. In particular, we focus on a new task which we term MIMDE: Multi-Insight Multi-Document Extraction.

MIMDE is a type of unsupervised text mining and summarisation task that involves a two-step process: first, extracting an unknown - but optimal - set of insights from a corpus of documents, and second, mapping these insights back to the source documents. Each insight is a piece of information that (a)~has utility to a human analyst with a specific goal, (b)~is specific, actionable, and context-aware, and (c)~may be shared and built upon across multiple documents. 
Examples of MIMDE tasks may include (i)~analysing customer feedback to identify specific product improvements, (ii)~synthesising patient experiences from medical records to enhance healthcare services, and (iii)~extracting key lessons from surveys containing open-ended responses. Using the last use case as an example, when analysing responses to the survey question \textit{``Do you believe the voting age should remain at 18 for UK Parliamentary elections? Explain why"}, a more traditional theme might be ``Taxes and vote eligibility", whereas an insight would be ``Anyone that pays taxes should be allowed to vote".

In this paper, we first contextualise MIMDE by reviewing previous work in this area before presenting a formal definition of these tasks. 
We then go on to make the following contributions:
\begin{enumerate}
    \item We introduce a pair of complementary datasets: the first features high-quality annotated data from humans responding to five survey questions, and the second offers synthetic versions generated by LLMs based on those human responses.
    \item We propose an evaluation framework for MIMDE tasks. As part of this, we present a methodology that evaluates the automated mapping of predicted and true insights.
    \item We evaluate a broad range of LLM models on both datasets using a naive brute-force approach to establish an initial baseline benchmark for performance on MIMDE tasks.
    \item By comparing performance across the two datasets introduced here, we assess the effectiveness of using synthetic data as a proxy for human-generated data in evaluating MIMDE tasks.
    \end{enumerate}

\section{Literature Review}
\label{sec:lit_review}
The ability of LLMs to analyse large volumes of unstructured text has led to its application in many routine processes \cite{kowalski2020improving,leelavathy2021public,hristova2022design,jagannathan2022application,alhelbawy2020nlp}. One key area of application that remains largely unexplored is the use of generative AI for unsupervised information extraction \textemdash where numerous documents are analysed without the exact prior knowledge of specific target information.
By enabling the extraction of information in an unsupervised manner, recent advancements in LLMs offer the potential to substantially reduce the resource and time requirements typically associated with analysing bespoke datasets \cite{huang2024text,viswanathan2023large}. 


\subsection{Information Extraction Tasks}
Traditional approaches 
employed to extract key information typically involve using keyword extraction \cite{garcia2008text,yih2007multi,quan2014unsupervised}, named-entity recognition (NER) \cite{tulkens2019unsupervised,weston2019named,wang2022automated}, graph-based approaches \cite{mihalcea2004textrank,sripada2005multi}, and leveraging large knowledge bases to build libraries that can then be used to classify documents \cite{tao2012unsupervised}.
Over time however a gradual shift has been observed in language-based tasks from classical NLP to deep learning methodologies and specifically, the use of generative AI and LLMs \cite{qin2024large}. 
LLMs are being increasingly exploited to review large volumes of unstructured data, including complex scientific literature and technical documents, to effectively extract structured knowledge \cite{dunn2022structured,dagdelen2024structured,gilardi2023chatgpt}. 

In a similar way, LLMs are also being used to ease the burden of classification and annotation tasks. More specifically within the domain of healthcare, studies are exploring the use of LLMs in conjunction with expert intervention for thematic coding of patient responses \cite{gamieldien2023advancing,meng2024exploring}. Thematic coding is typically used in clinical and qualitative research to extract critical insights from patient (or survey) responses followed by the process of labelling (or coding) these responses with the extracted insights \cite{braun2012thematic}. The process traditionally involved human-defined coding rules and typically the use of supervised machine learning to automate the annotation process \cite{rietz2020cody, goel2023llms}. 
The MIMDE task introduced in this paper is a fusion of unsupervised information extraction and classification, made possible by LLMs.

\subsection{LLM Performance and Reliability}
Despite promising developments in NLP, opinions on the use of LLMs in practice remain divided. Some studies indicate that LLMs can outperform crowd-workers in text annotation tasks \cite{gilardi2023chatgpt}, while others highlight the necessity of a human-in-the-loop approach \cite{dai2023llm}. In particular, the trustworthiness of LLMs is a topic of much debate given the emergent capabilities of these models and the complexity of tasks performed by LLMs \cite{sun2024trustllm}. This issue of trustworthiness is enhanced when dealing with business critical tasks or when analysing sensitive data, and there is a critical need for a thorough evaluation of LLMs to ensure their reliability and effectiveness in practice. 
Typically, platforms such as the \href{ https://huggingface.co/spaces/open-llm-leaderboard/open_llm_leaderboard }{Open LLM Leaderboard}, \href{https://crfm.stanford.edu/helm/}{HELM Leaderboards}, and \href{https://chat.lmsys.org/}{LMSYS Chatbot Arena} have been utilised to monitor the performance of LLMs on a range of reasoning, bias, and human preference tasks in an open and standardised manner. Here we present a comprehensive assessment of the performance of a range of state-of-the-art LLM models on this new task to rigorously examine their reliability and accuracy on two related datasets generated for the purpose of this evaluation.

\subsection{Synthetic Data for Model Evaluation}
Benchmark datasets are critical when evaluating the performance of LLMs in practice. In some instances, standardised datasets are readily available for evaluation purposes, such as the \href{https://www-nlpir.nist.gov/projects/duc/data.html}{DUC datasets} for summarisation tasks. However for some other tasks, there is a lack of evaluation data, in which case synthetic datasets serve as a valuable alternative \cite{liu2021question,joshi2024synthetic}.
Often a key motivation for generating and utilising synthetic datasets for model evaluation stems from the sensitive nature of the underlying data and the associated privacy concerns \cite{jordon2022synthetic}, particularly in regards to clinical data \cite{tang2023does}. Additionally, obtaining data can often be challenging due to the high costs associated with the data collection process \cite{srivastava2022beyond, reddy2019coqa, rajpurkar2016squad}. 
For example, LLMs are rigorously evaluated for their ability to maintain coherent conversations with users. To this effect, a large number of synthetic dialogue datasets have been generated to benchmark the performance of LLMs \cite{byrne2019taskmaster,abdullin2024synthetic,zhao2018zero}. Similarly, the CodeXGLUE synthetic dataset was proposed as a benchmark dataset to evaluate code generation in LLMs \cite{lu2021codexglue}.
\section{Methodology}
\label{sec:methods}

\subsection{MIMDE Defintion}
We formally define MIMDE tasks below.\\
\\
Given a set of documents \( D \):
$$  d_1, d_2,{\ldots},d_n \in D \;$$
and a set of insights \( I \):
$$ i_1, i_2,{\ldots},i_m \in I \;$$
Find a method which \text{Extracts} \( I \) from \( D \) and
$$\text{classifies} \; d_1,d_2,...,d_n \; \text{into} \; I_j \;$$
$$ \text{where } I_j \subseteq I \; \text{for} \; j \text{ in } 1:n $$

In the context of the datasets introduced in the following \hyperlink{subsection.3.2}{section} a document ($d_{i}$) refers to a single free-text response (either human-generated or synthetically-generated) to a specific survey question. An insight ($i_{i}$) as outlined in \hyperlink{section.1}{Section 1} is a specific, actionable and context-aware piece of information which has been extracted from a set of documents (D). 



\subsection{Data Collection and Generation} 
 
As outlined earlier, here we focus on MIMDE tasks within a specific domain \textemdash analysis of free-text survey responses.\footnote{Here our survey takes the form of a public consultation, which is a tool used by governments to survey public sentiment towards a policy (or policy change)\cite{lga_web}.}
We present two new complementary datasets based on the same set of five hypothetical questions (which can be found in Table  \ref{tab:survey}): one comprising human-generated responses, and another containing synthetic responses generated by state-of-the-art LLMs.

\subsubsection{Human Data}
Over 1,000 participants were recruited through \href{https://www.prolific.com}{Prolific} to participate in this study, with the survey administered via \href{https://www.qualtrics.com}{Qualtrics}. For each of the five questions, participants were asked to indicate their stance using multiple-choice options (e.g., agree/disagree/neither) and provide a free-form text response explaining their choice. The survey took approximately 11 minutes to complete, with participants compensated at an average rate of £11.47/hr.



To identify the insights within the human-generated responses and map responses to these insights, we followed a two-step process. 
First, we conducted three pilot surveys to test question wording and to collect an initial set of free-text responses to each of the questions. These responses were reviewed by the research team to prepare a list of expected insights in preparation of the full survey. In the final survey of 1,000 participants, after writing their free text, respondents were asked to label their responses with insights from this list and where appropriate, they could add their own insights. To limit biases in the final dataset, members of the research team reviewed a random sample of 200 responses for each question and updated the list of final insights. 

Second, 1,875 annotators were recruited through Prolific to each annotate eight random responses using the list of insights prepared in the previous stage. Each response in the dataset was thus annotated by three separate annotators. To ensure high quality and reliability of the final dataset, we mapped an insight to a response only when two out of the three annotators agreed on its presence. The annotation process achieved strong consensus \textemdash in all instances at least two annotators agreed on the presence (16\%) or absence (84\%) of each insight, with full agreement among all three annotators 81\% of the time. Table \ref{tab:survey} provides general details on the number of insights identified for each question.

\subsubsection{Synthetic Data}
To complement our human-generated data, we generated a synthetic dataset using several LLMs, such as GPT-4, GPT-3.5, Mistral-Large, and Llama2-70B. These LLMs were prompted to respond to the same five hypothetical consultation questions and their responses were streamlined by insights given as inputs.
More specifically, each prompt was injected with a randomly selected multiple choice response and 1--3 insights. The injected insights were randomly picked from a pre-determined set of insights derived by the research team from the human-generated data and the LLM was instructed to build a free-text response based on these insights. Thus producing a labelled response dataset for each question. To ensure diversity in the dataset, we introduce some stochasticity in the responses. We do this by creating unique prompts that incorporate the following additional features:

\begin{enumerate}
\item Each insight was paired with a randomly selected sub-insight to provide more specific guidance.
\item The LLM was instructed to adopt a unique personality, defined by a randomly assigned character trait (e.g., ``educated," ``passionate," ``skeptical"), sex, and age group.
\item We varied the temperature (0.9--1.35) and presence penalty (0.3--0.6) parameters for each LLM call to introduce additional variability in the responses.
\end{enumerate}

Furthermore, we used few-shot prompting by including three random examples from the dataset of human responses in each prompt, to guide the model's output. This approach allowed us to generate a diverse set of synthetic responses that mirrored the structure and content of our human-generated data while introducing controlled variations. The full version of both datasets as well as an example of the prompt used to generate our synthetic data is available upon request, and will be published publicly at a later date. 

\begin{table*}[t]
\centering
\resizebox{\textwidth}{!}{%
\renewcommand{\arraystretch}{3}
\begin{tabular}{p{12cm} | p{2cm} p{2.3cm} p{2.3cm} | p{2cm}  p{2.3cm} p{2.3cm}}
 \hline
\multirow{2}{*}{\centering \textbf{\LARGE Question}}   & \multicolumn{3}{l}{\centering \textbf{\LARGE Human Data}} & \multicolumn{3}{l}{\centering \textbf{\LARGE Synthetic Data}} \\
\cline{2-7}
  & {\Large \#Insights} & {\Large Insights/ Response} & {\Large \Large Words/ Response} & {\Large \#Insights} & {\Large Insights/ Response} & {\Large Words/ Response} \\
 \hline
{\Large To what extent do you agree or disagree with the changes to the highway code which states that: drivers and motorcyclists should switch off their mobile phone before setting off? (Strongly agree / agree / disagree / strongly disagree / Neither agree nor disagree) Please explain your answer.} & {\Large 12} & {\Large 1.3} & {\Large 167.3} & {\Large 11} & {\Large 2.0} & {\Large 239.4} \\
\hline
{\Large The UK government is planning to ban parking on the pavement everywhere in the UK. Do you think this change would result in more positive or negative impacts with regards to accessibility? Please explain.} & {\LARGE 13} & {\Large 1.3} & {\Large 168} & {\Large 14} & {\Large 2.0} & {\Large 223.7} \\
\hline
{\Large Please rank what you think the speed limit for inner-city (built-up) areas should be: 20MPH / 30MPH / 40MPH Please explain your answer.} & {\Large 15} & {\Large 1.3} & {\Large 139.5} & {\Large 14} & {\Large 2.0} & {\Large 221.1} \\
\hline
{\Large How would you describe your experience with mobile internet (e.g. internet on your cell phone) in the UK?} & {\Large 8} & {\Large 1.3} & {\Large 118.4} & {\Large 6} & {\Large 2.0} & {\Large 219.6} \\
\hline
{\Large For UK Parliamentary elections, the current voting age is 18. I think the voting age should be lowered to ... 16 / stay at 18 / be raised to 21. Please explain.} & {\Large 12} & {\Large 1.2} & {\Large 146.6} & {\Large 15} & {\Large 2.0} & {\Large 229.0} \\
\hline
\end{tabular}
}
    \caption{Dataset characteristics across five survey questions: comparing human and synthetic responses}
    \label{tab:survey}
\end{table*}

\subsection{Evaluation Framework}
We now present a methodology to evaluate the performance of MIMDE tasks. As outlined above, for each consultation question we have a list of true insights denoted by:
$$I_T:{i_{t_1}, i_{t_2},{\ldots},i_{t_m}}$$
and a list of LLM-extracted predicted insights denoted by:
$$I_P:{i_{p_1}, i_{p_2},{\ldots},i_{p_m}}$$

The alignment between $I_T$ and $I_P$ is measured using three metrics:
\vspace{-5pt}
\begin{itemize}\setlength{\itemsep}{-2pt}
    \item True Positive $(TP)$: instances where a true insight in $I_T$ matches at least one predicted insight in $I_P$;
    \item False Positive $(FP)$: instances where a predicted insight in $I_P$ does not match any of the true insights in $I_T$;
    \item False Negative $(FN)$: instances where a true insight in $I_T$ does not match any of the predicted insights in $I_P$;
\end{itemize}
\vspace{-5pt}

However, determining whether a predicted insight matches a true insight is not straightforward because insights can express the same core idea using different words, phrasings, or levels of detail. This is a common and much studied issue for tasks which require comparisons of short texts \cite{prakoso2021short}. To address this challenge, we created an additional dataset specifically designed to evaluate how various similarity metrics align with human judgment when comparing pairs of insights.

\subsubsection{Insight Similarity}

Our first research question in this work asks: What method best aligns to human preference for comparing the similarity of two insights? To evaluate similarity metrics, we created a benchmark dataset by manually mapping insights that exist within our two datasets to a set of synthetically generated insights. This dataset comprises 121 distinct true insights (identified as described in \hyperlink{subsection.3.2}{Section 3.2}) and 1,288 distinct predicted insights generated by LLM models for the questions in Table \ref{tab:survey}. Each row in the dataset pairs a predicted insight with a true insight for a given question, resulting in 9,378 total paired comparisons. Members of the research team then manually labeled these pairs, identifying 1,254 rows (13\%) as matches—cases where a predicted insight was judged equivalent to a true insight—with the remainder labeled as non-matches.

We examined how different similarity metrics compare to human perceptions across four distinct categories of metrics. The first category, distance-based metrics, quantify the difference between two text strings by considering various string operations and character patterns. Here, we selected the normalised Levenshtein distance, which compares strings through edit operations (deletions, insertions, and substitutions) while accounting for string lengths \cite{yujian2007normalized}, and the Jaro-Winkler distance, which has been found to perform well for lexically similar terms \cite{prasetya2018performance}.

The second category, keyword-based metrics, is similar to distance-based metrics but places less emphasis on word order and adds bonuses for direct unigram and bigram matches. From this category, we evaluated three versions of ROUGE, the BLEU metrics, the METEOR metric, and two versions of TF-IDF, which accounts for the relative occurrence of each unigram and bigram. 

Our third category comprises semantic metrics, which convert text pairs into numerical embeddings and compares their vector similarity, typically using cosine similarity. We tested several embedding approaches: OpenAI embeddings (\textit{text-3-large}), BLEURT, Distilbert embeddings, General Text Encoding embeddings, MXBAI embeddings, and the Universal Sentence Encoder from Google. Finally, we evaluated a set of pre-trained LLMs, which were prompted using a simple zero-shot prompt with chain-of-thought reasoning to judge similarity.

In total, we evaluated 17 similarity metrics across these categories using our insight-mapping dataset. Each metric generates a similarity score between 0 and 1 for each pair of insights. To determine optimal performance, we tested various thresholds for each metric where scores equal to or above the threshold were counted as matches, while scores below were counted as non-matches. For each metric, we identified the threshold that maximised the number of matches. 

\subsubsection{Metrics for Evaluating MIMDE}
 We evaluate performance on each question in our datasets at both the insight extraction level and the individual document categorisation level using four metrics. The precision score measures the proportion of predicted insights that match the true insight set: $\frac{TP}{(TP + FP)}$. Recall captures the proportion of true insights identified by the method: $\frac{TP}{(TP + FN)} $, and the weighted F1 score represents a balance between precision and recall: $ \frac{2*TP}{2*TP + FN + FP} $. Additionally, we introduce a relevancy rate to measure redundancy in the predicted insights: $\frac{TP}{MP}$, where $MP$ is the number of matched predictions between true and predicted insights. For each model, we calculate the average of these metrics across all five questions to obtain overall precision, recall, F1, and redundancy scores.

\subsection{Benchmark Implementation Approach}
To create an initial benchmark of MIMDE tasks, we tested the performance of 20 different state-of-the-art language models on our human-generated and synthetically-generated datasets. Our goal was twofold: First, assess how well state-of-the-art LLM models perform on an example MIMDE task to set a baseline for MIMDE performance and second, compare the performance of LLM models on human and synthetic data. 

To evaluate LLM performance, we implemented a naïve brute force approach consisting of two steps: first insights were extracted from the responses, then these extracted insights were mapped back to individual responses.
In the extraction step, we iteratively prompted each LLM with a set of instructions, the question, the multiple choice selection, and a batch of responses that fit within the model's context window. When responses exceeded the context window limit, we processed the question in chunks, repeating this process until the set of insights from the previous round could fit into a single prompt. This iterative approach enabled the consolidation of insights into a final unique list.
For the mapping step, we provided each LLM with both the list of extracted insights and a batch of responses, prompting it to associate relevant insights with each response. 

\subsection{Synthetic vs. Human Performance}
As part of this study we were particularly interested in the second research question: \textit{Does synthetically-generated data provide a good approximation of human-generated data for evaluating MIMDE tasks?} This is of particular interest as the creation of high quality human evaluation data for many MIMDE tasks can be both time- and resource-consuming. Thus, having a methodology that can generate a reliable proxy dataset would open up the possibility of generating large numbers of synthetic datasets for evaluation purposes \textemdash potentially accelerating the development and refinement of MIMDE and similar tasks. Our hypothesis is that the synthetic data introduced here should well approximate the data collected from humans. To compare performance on the novel datasets introduced here, we conducted a correlational analysis using the Pearson correlation coefficient to see if performance on the synthetic and human responses follow a positive relationship \textemdash that any change in the performance of an LLM on the human data is equivalent to a proportional change in the performance on the synthetic data. We were also particularly interested to see if the most performant models on the synthetically-generated data were also the most performant models on the human-generated data.
\section{Results}
\label{sec:results}

\begin{figure*}[!htbp]
\centering
\includegraphics[width=1\textwidth]{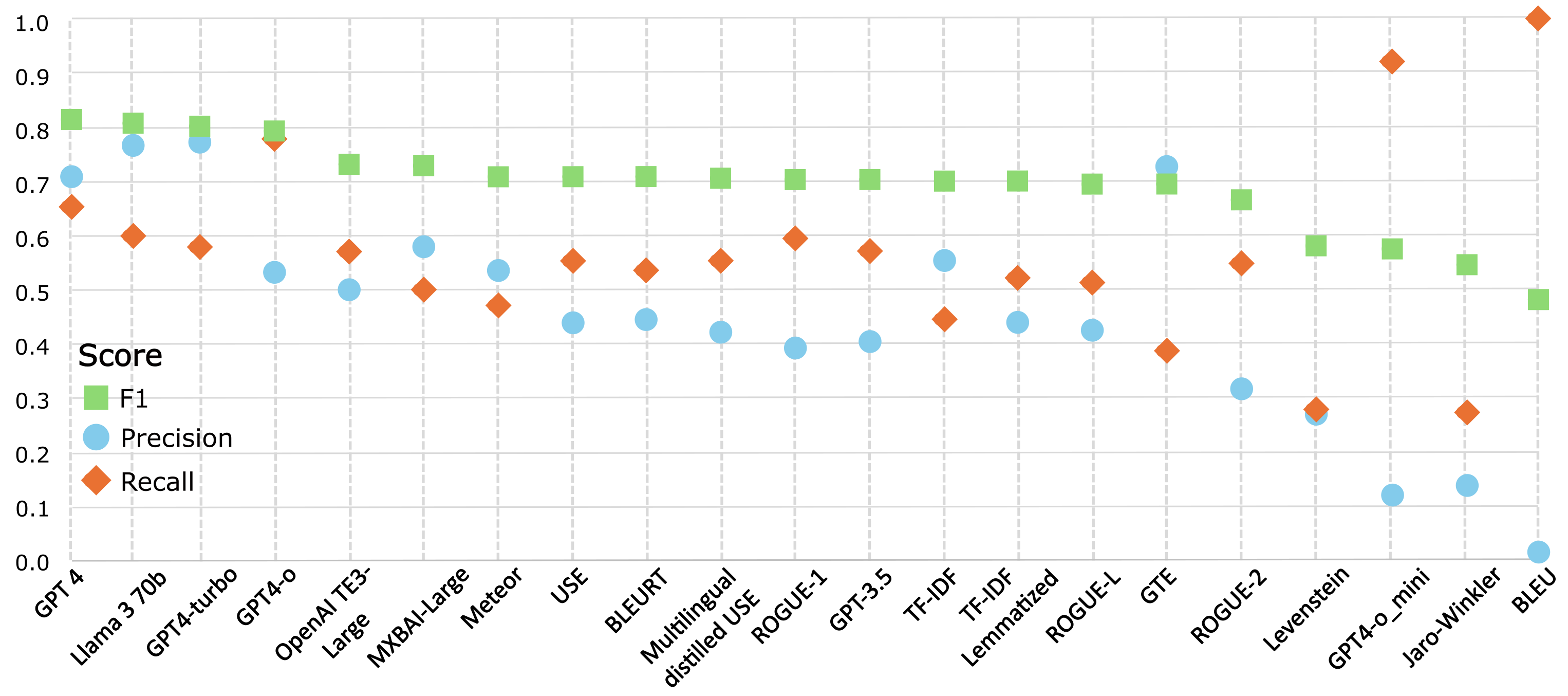}
\caption{\textbf{Evaluating Potential MIMDE Metrics} (\textit{Precision}) Measures how often true matches to human mapped insights are correctly identified. (\textit{Recall}) Measures how well the model identifies true positives from all the true positives in the dataset. (\textit{F1}) A weighted mean between Precision and Recall. }
\label{fig:fig1}
\end{figure*}

\subsection{Evaluating Similarity Metrics}
Figure \ref{fig:fig1} presents the F1, precision, and recall scores for all examined similarity metrics on the insight-mapping dataset, sorted by F1 score. Among the metrics tested, GPT-4's similarity judgments aligned most closely with human preferences, achieving a peak F1 score of 0.809.
It is worth mentioning here that examining recall or precision in isolation can be misleading for this task. For instance, if a model labels all pairs as matches (or all as non-matches), one of these metrics will automatically be 1, as demonstrated by the BLEU metric. Hence, F1 score provides a more balanced and realistic assessment of the performance of each metric in this context.

Overall, LLM and semantic metrics showed superior alignment with human preferences. Distance-based and keyword-based metrics performed poorly, struggling to identify matches when insights expressed similar meanings using different vocabulary. While this weakness in distance and keyword-based methods is expected given they were not designed to capture semantic meaning, it is noteworthy that state-of-the-art LLMs like GPT-4 and LLama-3-70B outperformed semantic metrics, despite embedding models often being designed to specifically capture and encode semantic similarity \cite{chandrasekaran2021evolution}.

\begin{table*}[t]
\centering
\scriptsize
\begin{tabular}{lcccc|lcccc}
& \multicolumn{8}{c}{\textbf{Insight level Performance}} \\
\toprule
& \multicolumn{4}{c}{\textbf{Human}} & \multicolumn{4}{c}{\textbf{Synthetic}} \\
\cmidrule(lr){2-5} \cmidrule(lr){6-9}
\textbf{Method} & \textbf{Recall} & \textbf{Precision} & \textbf{F1} & \textbf{Redundancy} & \textbf{Recall} & \textbf{Precision} & \textbf{F1} & \textbf{Redundancy} \\
\midrule
Cohere                 & 0.80 & 0.64 & 0.67 & 0.29 & 0.92 & 0.78 & 0.82 & 0.28 \\
Phi-3.5-MoE            & 0.76 & 0.65 & 0.67 & 0.38 & 0.86 & 0.88 & 0.86 & 0.20 \\
llama-3-70b-instruct   & 0.73 & 0.70 & 0.70 & 0.26 & 0.86 & 0.80 & 0.81 & 0.17 \\
llama-3-1.405B         & 0.72 & 0.66 & 0.67 & 0.41 & 0.93 & 0.85 & 0.87 & 0.35 \\
Phi-3.5-mini           & 0.71 & 0.73 & 0.70 & 0.36 & 0.85 & 0.90 & 0.86 & 0.19 \\
llama-3-1.8b-instruct  & 0.71 & 0.56 & 0.56 & 0.63 & 0.84 & 0.61 & 0.60 & 0.58 \\
Mistral-large          & 0.70 & 0.68 & 0.66 & 0.24 & 0.86 & 0.94 & 0.89 & 0.10 \\
llama-2-70b-chat       & 0.65 & 0.41 & 0.38 & 0.47 & 0.80 & 0.61 & 0.60 & 0.46 \\
gpt-4                  & 0.64 & 0.85 & 0.72 & 0.27 & 0.84 & 0.95 & 0.89 & 0.10 \\
gpt-4o                 & 0.64 & 0.76 & 0.68 & 0.34 & 0.83 & 0.94 & 0.87 & 0.11 \\
gemini-1.5-pro         & 0.62 & 0.82 & 0.69 & 0.25 & 0.85 & 1.00 & 0.91 & 0.17 \\
gpt-4o-mini            & 0.62 & 0.77 & 0.67 & 0.27 & 0.85 & 0.98 & 0.90 & 0.12 \\
Phi-3-medium           & 0.61 & 0.81 & 0.68 & 0.40 & 0.74 & 0.93 & 0.80 & 0.11 \\
llama-2-70b-chat       & 0.60 & 0.71 & 0.62 & 0.35 & 0.88 & 0.90 & 0.87 & 0.28 \\
gemini-1.5-flash       & 0.60 & 0.85 & 0.69 & 0.28 & 0.76 & 0.97 & 0.84 & 0.07 \\
llama-3-8b-instruct    & 0.59 & 0.83 & 0.67 & 0.29 & 0.77 & 0.92 & 0.83 & 0.18 \\
Mistral-small          & 0.59 & 0.81 & 0.66 & 0.22 & 0.74 & 0.90 & 0.80 & 0.07 \\
gemini-1.0-pro         & 0.57 & 0.92 & 0.69 & 0.35 & 0.76 & 0.92 & 0.81 & 0.17 \\
gpt-35-turbo-16k       & 0.56 & 0.74 & 0.63 & 0.29 & 0.81 & 0.91 & 0.83 & 0.26 \\
Phi-3-small-8k         & 0.46 & 0.76 & 0.56 & 0.25 & 0.68 & 0.95 & 0.77 & 0.22 \\
\bottomrule
\end{tabular}
\caption{Recall, Precision, F1, and Redundancy metrics for Human and Synthetic dataset at the insight level.}
\label{tab:insight_tab}
\end{table*}

\begin{table*}[t]
\centering
\scriptsize
\begin{tabular}{lccc|lccc}
 & \multicolumn{6}{c}{\textbf{Response level Performance}} \\
\toprule
 & \multicolumn{3}{c}{\textbf{Human}} & \multicolumn{3}{c}{\textbf{Synthetic}} \\
\cmidrule(lr){2-4} \cmidrule(lr){5-7}
\textbf{Method} & \textbf{Recall} & \textbf{Precision} & \textbf{F1} & \textbf{Recall} & \textbf{Precision} & \textbf{F1}  \\
\midrule
gpt-4	               & 0.46 & 0.45 & 0.45 & 0.21 & 0.32 & 0.25 \\ 
Phi-3-small-8k	       & 0.42 & 0.47 & 0.40 & 0.01 & 0.92 & 0.02 \\
gemini-1.5-flash-002   & 0.42 & 0.42 & 0.33 & 0.21 & 0.55 & 0.21 \\
gemini-1.0-pro-001	   & 0.41 & 0.57 & 0.33 & 0.22 & 0.42 & 0.19 \\
llama-3-70b-instruct   & 0.40 & 0.38 & 0.38 & 0.09 & 0.33 & 0.13 \\
llama-3.1-405B	       & 0.40 & 0.26 & 0.31 & 0.14 & 0.26 & 0.18 \\
gpt-35-turbo-16k	   & 0.39 & 0.42 & 0.40 & 0.18 & 0.27 & 0.21 \\
gpt-4o	               & 0.38 & 0.37 & 0.36 & 0.18 & 0.28 & 0.21 \\
llama-3-8b-instruct	   & 0.38 & 0.36 & 0.36 & 0.04 & 0.57 & 0.05 \\
gemini-1.5-pro-002	   & 0.36 & 0.33 & 0.30 & 0.29 & 0.28 & 0.28 \\
llama-2-70b-chat	   & 0.34 & 0.28 & 0.30 & 0.09 & 0.34 & 0.12 \\
gpt-4o-mini	           & 0.32 & 0.36 & 0.33 & 0.18 & 0.30 & 0.22 \\
Mistral-small	       & 0.28 & 0.41 & 0.27 & 0.15 & 0.27 & 0.19 \\
Cohere	               & 0.27 & 0.41 & 0.27 & 0.11 & 0.40 & 0.14 \\
llama-2-7b-chat	       & 0.26 & 0.25 & 0.25 & 0.14 & 0.25 & 0.17 \\
Phi-3-5-MoE	           & 0.24 & 0.28 & 0.24 & 0.02 & 0.68 & 0.03 \\
Phi-3-medium	       & 0.18 & 0.52 & 0.24 & 0.02 & 0.59 & 0.04 \\
llama-3-1-8b-instruct  & 0.15 & 0.24 & 0.18 & 0.11 & 0.28 & 0.13 \\
Mistral-large	       & 0.10 & 0.86 & 0.07 & 0.16 & 0.25 & 0.19 \\
Phi-3-5-mini	       & 0.02 & 0.77 & 0.04 & 0.19 & 0.40 & 0.21 \\
\bottomrule
\end{tabular}
\caption{Recall, Precision and F1 metrics for Human and Synthetic dataset at the response level}
\label{tab:response_tab}
\end{table*}

\subsection{Benchmarking MIMDE Performance} 
To benchmark performance on MIMDE, we evaluated 20 different state-of-the-art LLMs on both the human-generated and synthetic datasets. To automatically map predicted and true insights for evaluation, we used GPT-4, which showed optimal performance among similarity metrics (Figure \ref{fig:fig1}). Tables \ref{tab:insight_tab} and \ref{tab:response_tab} present each model's performance at both the insight and document levels, sorted by recall scores on the human dataset to highlight each LLMs' ability to extract all existing insights.

At the insight level (Table \ref{tab:insight_tab}), of the LLMs tested, Cohere Command R plus achieved the highest recall with a score of 0.80 on the human dataset, accurately identifying 80\% of true insights in the data. GPT-4 achieved the highest F1 score (0.72) on this dataset. Although GPT-4 identified 16\% fewer insights than Cohere, it was more precise in its predictions, generating fewer inaccurate or superfluous insights. On the synthetic dataset, we observed different performance patterns: LLama 3.1 405B demonstrated the highest recall with a score of 0.93, while Gemini 1.5 pro achieved the highest F1 score at 0.91.

Performance on the synthetic dataset consistently exceeded that of the human dataset by 0.1-0.2 \textemdash this is somewhat expected since the synthetic data was generated using controlled prompts, leading to more consistent patterns of expression compared to the natural diversity in human responses. The Mistral models demonstrated strong performance across both datasets in terms of redundancy rate (here lower scores indicate better performance), producing more concise outputs. In contrast, LLama and Phi models struggled with redundancy, often generating multiple variations of the same insight.
We observed a consistent precision-recall trade-off across models. Notably, even lower-performing models maintained relatively good precision in their predictions, despite capturing fewer insights from the datasets. This suggests that while these models were unable to capture as many insights, the ones they did identify were generally accurate.

Table \ref{tab:response_tab} presents the results of the document-level analysis, showing precision, recall, and F1 scores averaged across all documents, sorted by recall.\footnote{Redundancy score is not relevant at the document level.} The task of mapping predicted insights back to individual documents proved considerably more challenging for the LLMs. This makes intuitive sense as whenever an insight was missed in the initial extraction phase, it could not be correctly mapped to the documents, leading to compounded errors and lower overall scores.

At the document level, GPT-4 led performance on the human dataset with a recall score of 0.46 and F1 score of 0.45, while Gemini-1.5-Pro achieved the highest scores on the synthetic dataset with a recall score of 0.29 and F1 score of 0.28. Interestingly, scores on the synthetic dataset were consistently lower across all models and metrics, suggesting greater difficulty in insight-to-document mapping for our synthetic data than our human data. This is despite our expectation that the highly tailored synthetic responses would provide an easier task for the LLMs to analyse. We discuss the potential reasons for this discrepancy in Section \ref{sec:discussion}.

\subsection{Correlation Analysis of Synthetic and Human Data}
\begin{figure*}[!htbp]
\centering
\includegraphics[width=1\textwidth]{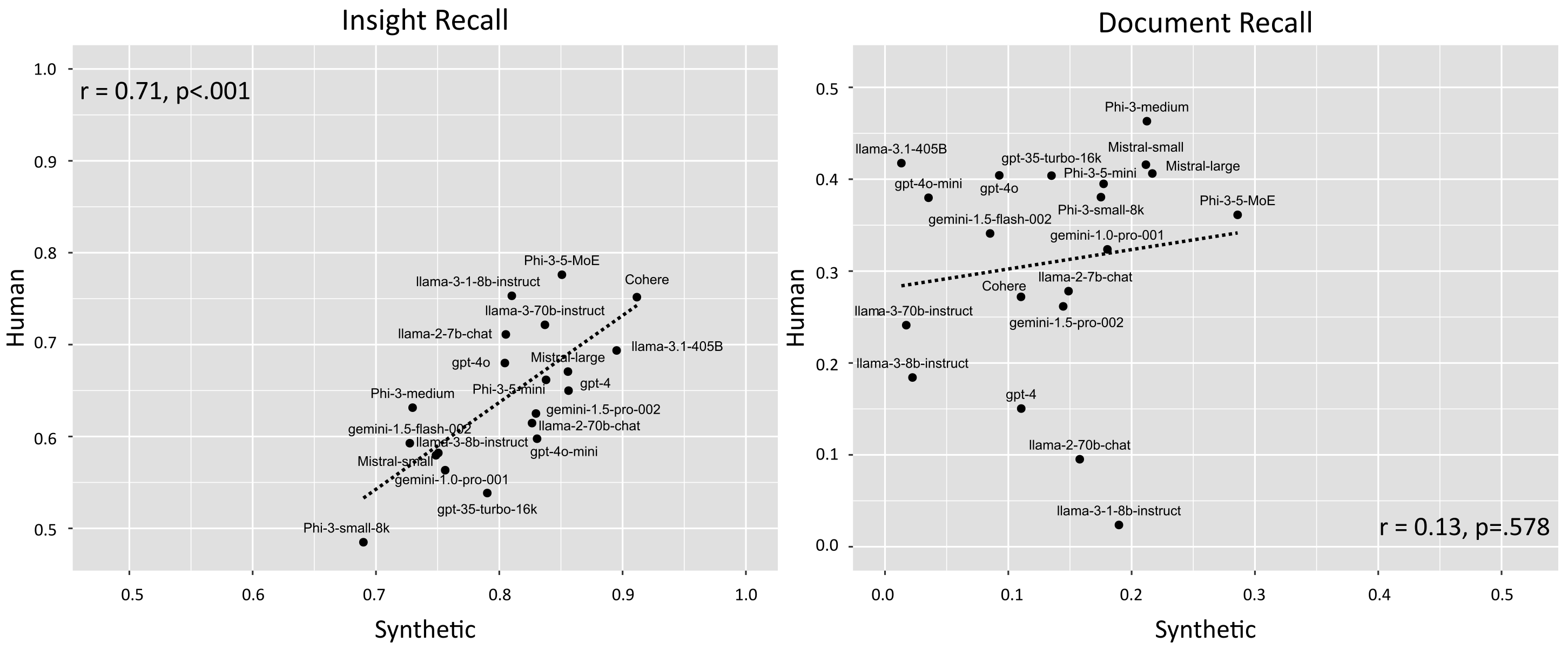}
\caption{\textbf{Relationship between Synthetic and Human data performance} (\textit{Recall}) Measures how often true insights have been correctly identified.}
\label{fig:fig_synth_hum}
\end{figure*}

Next we examined whether synthetic data could effectively approximate human data for MIMDE evaluation through a correlational analysis of LLM performance across both datasets. Figure \ref{fig:fig_synth_hum} illustrates these correlations for recall scores at both insight and document levels, based on the performance reported in Tables \ref{tab:insight_tab} and \ref{tab:response_tab}.

At the insight level, shown on the left side of Figure \ref{fig:fig_synth_hum}, we observed a strong positive relationship between recall scores across datasets (correlation = 0.71, p-value $< 0.001$). As shown in Table \ref{tab:correlation}, this strong correlation holds consistently across other reported metrics (F1, Precision, and redundancy rate). These findings suggest that for insight extraction tasks, relative performance of models remains consistent whether tested on synthetic or human-generated texts—models performing well on one dataset tend to perform well on the other.

However at the document level, shown on the right side of Figure \ref{fig:fig_synth_hum}, we found little to no relationship between the performance of LLMs across datasets (correlation = 0.13, p-value$ > 0.1$). This finding contradicts our hypothesis and suggests that our synthetic data, at least at the document level, may not serve as a reliable proxy for evaluating MIMDE performance on human-generated responses. How well an LLM assigns insights to documents on the synthetic data, is not indicative of how well that LLM will assign insights to human written responses.
We explore potential explanations for this discrepancy in the following discussion section.

\begin{table}[t]
\centering
\scriptsize
\begin{tabular}{lcc|lcc}
 & \multicolumn{2}{c}{\textbf{Insight}} & \multicolumn{2}{c}{\textbf{Document}} \\
\cmidrule(lr){2-3} \cmidrule(lr){4-5}
\textbf{Metric} & \textbf{Correlation} & \textbf{p-value} & \textbf{correlation} & \textbf{p-value} \\
\midrule
F1 & 0.796 & $<0.001$ & 0.001 & 0.996 \\
Recall & 0.712 & $<0.001$ & 0.132 & 0.578 \\
Precision & 0.864 & $<0.001$ & 0.065 & 0.786 \\
Redundancy & 0.797 & $<0.001$ & -- & -- \\
\bottomrule
\end{tabular}
\caption{Correlation Analysis of Human vs. Synthetic data}
\label{tab:correlation}
\end{table}


\section{Discussion}
\label{sec:discussion}

This work set out a definition of a group of natural language tasks brought about by the advent of LLMs which allow for the extraction and identification of a set of unknown insights from a large set of documents. We refer to these as MIMDE tasks and provided a framework to evaluate the performance of different language models on MIMDE, an area which has previously been dominated by bespoke solutions \cite{soldaini2016quickumls,verma2017evaluation,florescu2017positionrank}. To achieve this, we first introduced a human-generated dataset alongside a paired synthetic version of the dataset for a specific type of MIMDE task \textemdash analysing free text survey responses. We then evaluated 20 state-of-the-art LLMs on each of these datasets, creating a baseline benchmark of performance on MIMDE. Finally, we investigated the extent to which the synthetic data we generated offers a good approximation of human data for evaluating LLMs on MIMDE tasks. 

A critical component of MIMDE evaluation is determining the optimal method for comparing predicted and true insights. In evaluating different methods for comparing predicted and true insights, a crucial aspect of MIMDE assessment, we found that state-of-the-art LLMs outperformed traditional approaches including distance metrics, keyword metrics, and semantic metrics when compared to human judgment. However, given that even the best automatic evaluation metrics achieved moderate results (F1=0.809), we recommend manual mapping between predicted insights and ground truth labels for applications requiring high-confidence evaluations.

We next examined whether synthetic data could serve as a reliable proxy for human data in evaluating MIMDE performance, given the significant cost differences in data generation. The human-generated dataset, including response creation and insight annotation by three annotators for five questions, cost £7,771.83 to produce. In contrast, generating a comparable dataset using LLMs cost only £442.87, about 1/20th of the cost of the human dataset. Our analysis of these datasets revealed mixed results across the two components of MIMDE.

At the insight level, performance on synthetic data strongly correlated with models' ability to extract insights from human-generated documents. However, at the document level, we found no relationship between synthetic and human data performance in mapping insights back to source documents. In addition to this, we found that it was more challenging to discover the insights at the document level for synthetic data compared with human data. This lower performance and lack of correlation likely stems from two key characteristics of our synthetic data: first, synthetic responses were typically longer, containing 0.7 more insights on average than human responses; second, while our synthetic data generation process involved prompting LLMs with specific insights to include, the models may not have incorporated all prompted insights into their responses, or may have expressed them in ways that made them difficult to detect. This highlights a limitation of our synthetic data generation approach and suggests that future work should include an additional verification phase to confirm the presence of intended insights, rather than relying solely on generation prompts for ground truth labeling.

There are important limitations to consider in our study. First, due to the high costs of generating human-annotated data, we could only explore MIMDE in the context of survey analysis. While this provides valuable insights, examining MIMDE across other domains—such as medical records or probation reports—would offer a more comprehensive understanding of LLM performance. 
Another challenge in evaluating MIMDE is the assumption of an objective ``ground truth" set of insights for each document. This assumption, while necessary for evaluation, represents a key limitation of MIMDE and similar tasks. The subjective nature of insight identification is evident in our survey dataset, where all three annotators agreed on the exact same set of insights for only 23\% of responses. To address this challenge, we adopted a majority voting approach, considering an insight present or absent when at least two of the three annotators agreed. This low initial agreement rate underscores why developing robust evaluation methods was a central focus of our work.
Lastly, to ensure fair comparisons, we used identical prompts across all LLMs. While this standardization was necessary for a comparative analysis, it likely understates the true capabilities of many models. Different LLMs often perform better with specific prompt structures, lengths, and styles, suggesting that model-specific prompt engineering could yield improved results.

Beyond addressing these limitations, several promising research directions emerge. The synthetic data generation process could be enhanced through advanced prompting techniques \cite{guo2024generative}, more sophisticated personas \cite{chan2024scaling}, and systematic verification of prompted insights in generated responses. To expand upon our correlation analysis, further work could attempt to predict the performance of LLMs on MIMDE tasks in real world applications from performance on synthetic data. This could involve testing a broader range of models and generating different variants of synthetic data to build a more comprehensive understanding of the relationship between synthetic and human-data performance.
Additionally, future work should examine potential biases in MIMDE performance. This includes investigating whether LLMs exhibit systematic differences in insight extraction across demographic groups, such as varying accuracy when analyzing responses from different age groups, cultural backgrounds, or language proficiency levels. Understanding these biases is crucial for ensuring MIMDE tasks are analysed equitably across diverse populations and use cases. 

\paragraph*{Funding Statement.} This work was supported by Towards Turing 2.0 under the EPSRC Grant EP/W037211/1 and The Alan Turing Institute.

\paragraph*{Data Availability Statement.} 
The code and datasets are available upon request and will be published publicly at a later date. 

\bibliographystyle{apalike}
\bibliography{main}


\end{document}